\documentclass[runningheads]{llncs}

\usepackage{booktabs}
\usepackage[T1]{fontenc}
\def\doi#1{\href{https://doi.org/\detokenize{#1}}{\url{https://doi.org/\detokenize{#1}}}}
\usepackage{graphicx}
\usepackage{listings}
\graphicspath{{./figs/}}
\DeclareGraphicsExtensions{.png,.pdf}

\begin{document}
%

\title{An Investigation on Word Embedding Offset Clustering 
as Relationship Classification}
%
%
%

\author{
	Didier Gohourou 
	\and
	Kazuhiro Kuwabara
} 
\authorrunning{D. Gohourou, K. Kuwabara}
\institute{
	Ritsumeikan University
}

\maketitle              

\begin{abstract}
Vector representations obtained from word embedding are the source of many 
groundbreaking advances in natural language processing. They yield word 
representations that are capable of capturing semantics and analogies of words 
within a text corpus.
This study is an investigation in an attempt to elicit a vector representation 
of relationships between pairs of word vectors. We use six pooling strategies 
to represent vector relationships. Different types of clustering models are
applied to analyze which one correctly groups relationship types.  
Subtraction pooling coupled with a centroid based clustering mechanism shows  
better performances in our experimental setup.
This work aims to provide directions for a word embedding based
unsupervised method to identify the nature of a relationship represented by a 
pair of words.
\end{abstract}
\section{Introduction}
The study of pattern recognition in natural language processing requires a 
numerical representation of text. This is achieved by computing a vector 
representation for the words that compose text corpora. Words were originally 
represented using a one-hot encoding vector representation. The approach had 
shortcomings including sparsed and high dimensional vectors that are unable to 
capture the contextual meaning of words. Proposed by Bengio et al, word 
embedding is a neural language model that addresses the \textit{curse of 
dimensionality}.  The proposed model also provides a distributed representation 
of words, aligned with linguistic principles such as the context-dependent 
nature of meaning. Different implementations have spawned from the proposition
including word2vec~\cite{Mikolov2013a,Le2014}, GloVe~\cite{Pennington2014}, 
and fastText~\cite{Bojanowski2017}. Those word representation models are the 
backbone of many natural language processing (NLP) tasks including named entity 
recognition, sentiment analysis, and machine translation, which led to 
state-of-the-art breakthroughs in those respective NLP fields. 

Word embedding demonstrated additional properties, such as the ability to 
capture syntactic and semantic regularities in language~\cite{Mikolov2013b}.
Building from this insight, can we obtain vectors that can capture the type of 
relationship between word embedding? Can those relationship representations be 
effectively grouped with clustering models? This study is an attempt 
to answer those questions. It explores different \textit{pooling} approaches 
to obtain a vector that represents the relationship between word vectors based 
on their embedding. 
The study also uses different clustering models in an attempt to score the 
ability of the relationship vectors to be grouped. Answering those 
problems will point toward \textit{an unsupervised methodology to classify 
relationships between words}.  Our contributions can be emphasized as follows:
\begin{itemize}
  \item Explore for a word embedding representation of the 
    relationship for a pair of words.
  \item Analyze the clustering of those relationship vectors.
  \item Cue toward an unsupervised classification mechanism of relationships
    between words.
\end{itemize}

The rest of the study unfolds by first discussing related works including the 
clustering of word vectors and the elicitation of regularities in word vector 
space. Then present the methodologies adopted for word embedding relationship 
representation and the descriptions of the families of clustering algorithms 
selected. An experiment section follows to detail the data set and specific 
cluster algorithms used. A discussion analyzes the results and offers directions 
to apply the findings, before concluding the study.

\section{Related Work}

Vector-space word representations learned by continuous space language models 
such as word2vec models have demonstrated the ability to capture syntactic 
and semantic regularities in language~\cite{Mikolov2013a}. In their study 
Mikolov et al. created a set of analogy questions in the form 
\textit{"a is to b as c is to ..."}. To answer the analogy question, 
$y = x_b - x_a + x_c$ is computed. Where $x_a$, $x_b$, and $x_c$ are the 
embedding vectors for the words $a$, $b$, and $c$ respectively. $y$ is the 
continuous space representation of the word expected as the answer. Because $y$ 
is not likely to represent an existing word, the answer is considered to be the 
word with the embedding that has the greatest cosine similarity.
The results showed better performance than prior methodologies on the 
SemEval-2012 Task 2: Measuring Relation Similarity.
The more recent transformer-based language models have been applied to 
solve abstract analogy problems~\cite{Ushio2021}. After pointing out the lack
of attention to recognizing analogies in NLP, the study emphasized 
the strength and limitation of various models on psychometric analogy data sets 
from educational problems. The transformer models used in the study included 
BERT, GPT-2, RoBERTa, GPT-3. Their performances were compared against 
word embedding models, perplexity-based point-wise mutual information, and 
random answers. GPT-2 and RoBERTa were the top performer, while BERT performed
worst than word embedding.

The clustering of word vectors has been used to analyze how well an 
embedding model's word vectors can be grouped by topic or other
taxonomies. It has been applied to various domain-specific studies including
business~\cite{Gohourou2017} and medicine~\cite{Suarez-Paniagua2015}.
Zhang et al.~\cite{Zhang2018} demonstrated that incorporating word embedding 
with a kernel-based k-mean clustering method provides superior performances 
within a topic extraction pipeline. In their study, word2vec was used to extract
features from bibliometric data set, claiming the advantage to skip human 
feature engineering. A specifically designed k-mean clustering model is used 
for topic extraction. The clustering methodology is a polynomial kernel function
integrated into a cosine similarity-based k-mean clustering.
The performance of the proposed model was compared to different models used for 
topic extraction including a standard k-mean algorithm, principal component 
analysis, and a fuzzy c-mean algorithm.
Closer to our work, an exploration of the word-class distribution in 
word vector spaces~\cite{Sasano2020} investigated how well distribution models
including Centroid-based, Gaussian Model, Gaussian Mixture Model, k-Nearest 
Neighbor, Support Vector Machine, and OffSet, can estimate the likelihood of a 
word to belong to a class. The study experimented with pre-trained vectors from 
Glove, and classes from the WordNet taxonomy. 

While analogy tasks try to find the best word to complete a pair to be similar 
to another, our problem focuses on probing for a vector representation of 
word-pair relationships, so that they are efficiently grouped by clustering 
models. This by extension call for investigating the performances of 
clustering models for this task.

\section{Methodology}

\subsection{Relation Vectors}

To obtain a \textit{relationship vector} from 
vector representations of words, we experiment with different \textit{pooling} 
strategies. Here we define by pooling the mechanism by which we reduce a set of
vectors representing different words, into a single one. The obtained vector 
will represent the relationship between the set of pooled vectors.
Our first pooling strategy is to use the subtraction operator.
. This strategy is derived from the linguistic regularity 
observation such as $king - man + woman = queen$. We can deduce 
$king - man = queen - woman$, where we consider the subtraction, the operator 
that gives a representation of the type of relationship between word vectors 
on both sides of the equation. Thus if $v_r$ is the relation vector representing
the relation between the word vectors $v_s$ and $v_o$, we have:
\begin{equation}
  v_r = v_s - v_o
\end{equation}
The second strategy is to apply the absolute value function to each component 
of the vector resulting from the subtraction.
\begin{equation}
  v_r = \langle|v_{o_1} - v_{s_1}|, |v_{o_2} - v_{s_2}|, ..., 
        |v_{o_n} - v_{s_n}|\rangle
\end{equation}
Where $v_{o_i}$ and $v_{s_i}$ are respectively the $i^{th}$ dimensional 
coordinate of $v_{o}$ and $v_{s}$.
The third consist of adding the two vectors involved in the relationship. 
\begin{equation}
  v_r = v_s + v_o
\end{equation}

The fourth pooling strategy constitutes a vector with the coordinate obtained 
by taking the minimum of each dimensional coordinate of the word vectors 
involved in the relationship. 
\begin{equation}
  v_r = \langle min(v_{o_1}, v_{s_1}), min(v_{o_2}, v_{s_2}), ..., 
        min(v_{o_n}, v_{s_n})\rangle
\end{equation}

Conversely, the fifth pooling strategy, named max pooling, consists of creating 
the relationship vector using the maximum of each dimensional coordinate. 
\begin{equation}
  v_r = \langle max(v_{o_1}, v_{s_1}), max(v_{o_2}, v_{s_2}), ..., 
        max(v_{o_n}, v_{s_n})\rangle
\end{equation}

The last pooling strategy we use in our exploration is the average pooling.
\begin{equation}
  v_r = \langle (v_{o_1} + v_{s_1})/2, (v_{o_2} + v_{s_2})/2, ..., 
        (v_{o_n} + v_{s_n})/2 \rangle
\end{equation}

Table~\ref{tab:pooling} gives the summary of the pooling strategies considered
is this study.

\begin{table}[ht]
\begin{center}
\caption{Pooling mechanisms for relationship vector representation.}
\label{tab:pooling}
\begin{tabular}{l l}
\toprule
Name & Formula \\
\midrule
Substraction & $ v_r = v_1 - v_2 $ \\
Substraction absolute value & $|v_1 - v_2| $ \\
Addition & $v_r = v_1 + v_2 $ \\
Minimum & $v_r = \langle min(v_{o_1}, v_{s_1}), min(v_{o_2}, v_{s_2}), ..., 
        min(v_{o_n}, v_{s_n})\rangle $ \\
Maximum & $v_r = \langle max(v_{o_1}, v_{s_1}), max(v_{o_2}, v_{s_2}), ..., 
        max(v_{o_n}, v_{s_n})\rangle $ \\
Mean & $v_r = \langle (v_{o_1} + v_{s_1})/2, (v_{o_2} + v_{s_2})/2, ..., 
        (v_{o_n} + v_{s_n})/2 \rangle  $ \\

\bottomrule
\end{tabular}
\end{center}
\end{table}

\subsection{Clustering}

Clustering is an active research field, that consist of grouping similar objects
into sets. It can be achieved with various algorithms that differ in mechanisms 
employed to constitute a cluster. Although more than seventy clustering 
models that can be classified into nearly twenty categories are commonly 
used~\cite{Xu2015}, we experiment with four types of the most widely used 
clustering mechanisms, including centroid-based, hierarchical-based, 
distribution-based and density-based.

\subsubsection{Centroid-based clustering}
represents their cluster based on a central vector that is usually randomly 
selected and then optimized. k-mean~\cite{Lloyd1982} and its variants including 
k-mean++~\cite{Bradley1997} and k-medians are part of this clustering category.

\subsubsection{Hierarchical clustering}
aims to build a hierarchy of clusters, either bottom-up (agglomerative) or 
top-down (divisive). The agglomerative approach starts by considering each data
point as a cluster. Pairs of clusters are then gradually merged, moving up
the hierarchy. Conversely, the divisive approach, start by considering all data 
points as one cluster. Clusters are then progressively split moving down the 
hierarchy. A measure of dissimilarity is used to determine which clusters 
should be merged or split. The dissimilarity is measured with a distance 
metric and a linkage criterion.

\subsubsection{Distribution-based clustering}
assumes the data points follow a statistical distribution. While distribution
based clustering models can capture statistical relationships within attributes 
of the data point, they work efficiently when the data distribution is known and
the model parameters are set accordingly.

\subsubsection{Density-based clustering} forms clusters by grouping data points 
from dense areas into clusters. Such clustering algorithm allows 
arbitrary-shaped but they might not be precise with data sets of varying 
densities. Furthermore, outlier data points are considered noise. 
Common density-based clustering includes Density-Based Spatial Clustering of 
Application with Noise (DBSCAN)~\cite{Ling1972}, Ordering points to identify 
the clustering structure (OPTICS), and Mean-shift.

\section{Experiment}
This section describes the data set used and selected algorithms used for 
experimentation. The experiments are conducted with the clustering model 
implementation of the Scikit-learn~\cite{Pedregosa2011} and 
Scipy~\cite{Virtanen2020} python packages.

We use the adjusted random score to evaluate the clustering accuracy of each 
setting.

\subsection{Datasets}

We use GloVe pre-trained word vectors~\footnote{
https://nlp.stanford.edu/projects/glove} trained on a corpus composed of the 
2014's Wikipedia dump and the 5th edition of the English Gigaword. The 
training corpus has six billion tokens, four hundred thousand unique words, and 
is lower-cased.  Our experiments use the 100 dimensions pre-trained vectors 
version.


The word pairs are drawn from the word pairs analogy data set of Mikolov et al.
~\cite{Mikolov2013b}. It contains 14 categories of word pairs for analogies 
tasks. Table~\ref{tab:questions} provides a sample of the data set word pairs 
for two categories.

\begin{table}[ht]
\begin{center}
\caption{Sample of the word pairs dataset.}
\label{tab:questions}
\begin{tabular}{cc}
\toprule
Countries' capital & Currencies \\
\midrule
baghdad - iraq & japan - yen  \\
bangkok - thailand & korea - won  \\
beijing - china & latvia - lats \\
berlin - germany & lithuania - litas \\
bern - switzerland& macedonia - denar \\
\bottomrule
\end{tabular}
\end{center}
\end{table}

\begin{figure}[htp]
  \centering
  \includegraphics[width=\textwidth]{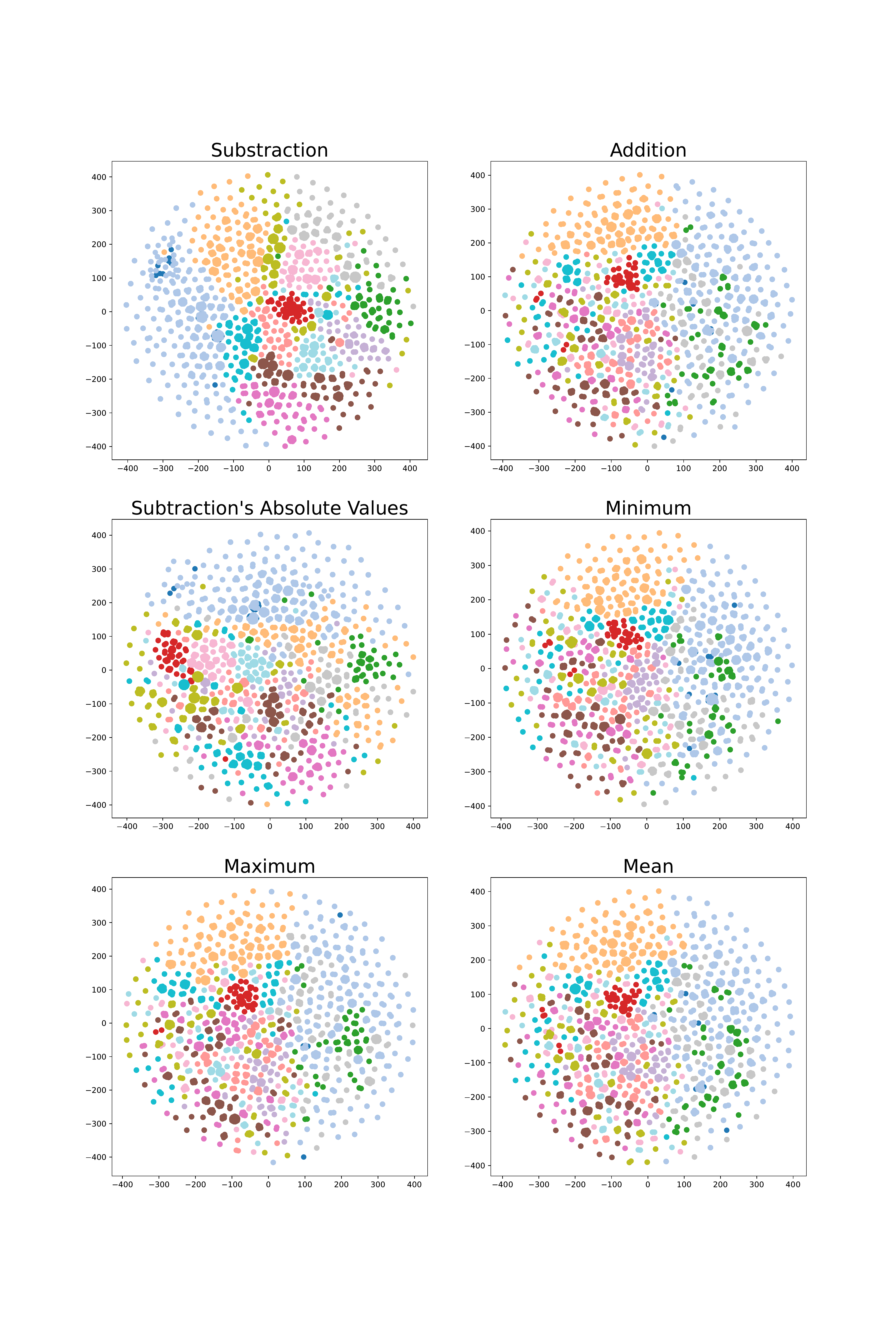}
  \caption{TSNE 2D projections of relation vectors for different pooling strategy
  }
  \label{fig:tsne_2d}
\end{figure}

For each word pair, their relation vector is obtained following the different
pooling strategies. Figure~\ref{fig:tsne_2d} gives a 2D projection overview of 
the relationship vectors for each pooling strategy.

\subsection{K-mean}
K-mean is not only the most used clustering mechanism but probably the most 
well-known clustering algorithm overall, because of its simple approach.
The basic running steps of k-mean are the following:
\begin{enumerate}
 \item Define the number of clusters and randomly initialize their center points.
 \item Each data point is added to the cluster of its closest center point.
 \item Center points are recomputed, using the \textit{mean} of all vectors in 
       a given cluster.
 \item The two previous steps are repeated, until the center points respectively
       converge.
\end{enumerate}
In addition to being simple, k-mean has a linear run time complexity $O(n)$. 
Some drawbacks are the need to specify the number of clusters, and the 
randomized initialization can provide different clustering results on 
different algorithms run.
Table~\ref{tab:kmean_score} provides the adjusted random scores for K-mean 
clustering.

\begin{table}[ht]
\begin{center}
  \caption{Adjusted random scores for K-mean clustering on the pooling
    strategies.}
  \label{tab:kmean_score}
  \begin{tabular}{lrrrrrr}
\toprule
 & $X_{subs} $ & $X_{add} $ & $X_{abs} $ & $X_{min} $ & $X_{max} $ & $X_{mean} $ \\
\midrule
kmean & \bfseries 0.792549 & 0.363478 & 0.296212 & 0.334140 & 0.313051 & 0.363478 \\
\bottomrule
\end{tabular}
 
\end{center}
\end{table}

\subsection{Gaussian mixture}
The Gaussian Mixture Model is a widely used clustering model of this type. It 
assumes that the point in the data set follows a gaussian distribution. The 
model can use two parameters such as the mean and the standard deviation to 
describe the shape of the clusters. The parameters are found using the 
Expectation-Maximization optimization algorithm. The Gaussian Mixture Model
clustering works as follows:
\begin{enumerate}
 \item Randomly initialize the Gaussian 
    distribution parameters for each cluster for the selected number of clusters.
 \item Compute the probability of each data point to belong to a cluster, 
    given the Gaussian distribution.
 \item Update the set of parameters for the Gaussian distributions, based
    on the data point probabilities. The new parameters are computed, maximizing
    the probability of data points within the cluster.
 \item The last two steps are iteratively repeated until the clusters' centers 
    respectively converge.
\end{enumerate}

The Gaussian mixture can be seen as a more flexible k-mean. Instead of assuming
clusters with a circular shape such as k-mean, Gaussian mixture can shape 
ellipse-like clusters.

Table~\ref{tab:gmm_score} provides the adjusted random score for the Gaussian
mixture model clustering. 
\begin{table}[ht]
\begin{center}
  \caption{The adjusted random score of the Gaussian Mixture Model for 
  \label{tab:gmm_score}
      different pooling strategies.}
  \begin{tabular}{lrrrrrr}
\toprule
 & $X_{subs} $ & $X_{add} $ & $X_{abs} $ & $X_{min} $ & $X_{max} $ & $X_{mean} $ \\
\midrule
gmm & \bfseries 0.854000 & 0.293000 & 0.351000 & 0.330000 & 0.228000 & 0.293000 \\
\bottomrule
\end{tabular}
 
\end{center}
\end{table}

\subsection{Agglomerative Clustering}
The hierarchical agglomerative clustering model can be described as follows:
\begin{enumerate}
 \item Every point in the data set is initially considered a cluster.
 \item Using a distance metric, clusters are jointly merged by pairs of the 
    closest to one another. 
 \item The previous step is repeated until a unique cluster is formed. The 
    clustering structure is then defined by choosing when to stop the clusters
    combination.
\end{enumerate}
The dissimilarity is measured with a distance metric and a linkage criterion.
We test various configurations of distance metrics as (dis)similarity measures,
and linkage criterion. The distance metrics used include euclidean distance, 
cosine similarity, manhattan, l1, and l2. The linkage criterion is the strategy 
used to merge clusters at different time steps. We experiment with the following
linkage criterion: 
\begin{itemize} 
  \item ward: minimize the variance of the clusters being merged,
  \item average: merge two clusters with the minimal average distances of each 
    observation,
  \item complete/maximum: merges two clusters with the smallest maximum distances
    between all observations of the two sets,
  \item single: merges two clusters with the smallest minimum distances between
    all observations of the two sets.
\end{itemize} 

The adjusted random scores of different configurations for the linkage and 
distance metric parameters are available in table~\ref{tab:agg_score}.

\begin{table}[ht]
\begin{center}
  \caption{The adjusted random score for different configurations of distance 
    and linkage parameters of agglomerative clustering, for different pooling 
    strategies}
  \label{tab:agg_score}
  \begin{tabular}{lrrrrrr}
\toprule
 & $X_{subs} $ & $X_{add} $ & $X_{abs} $ & $X_{min} $ & $X_{max} $ & $X_{mean} $ \\
\midrule
(ward, euclidean) & \bfseries 0.682373 & 0.302749 & 0.342485 & 0.317647 & 0.283547 & 0.302749 \\
(single, euclidean) & 0.006886 & \bfseries 0.007961 & 0.006723 & 0.004333 & 0.006045 & \bfseries 0.007961 \\
(complete, euclidean) & \bfseries 0.502632 & 0.296154 & 0.081222 & 0.167146 & 0.299644 & 0.296154 \\
(average, euclidean) & 0.022881 & \bfseries 0.293066 & 0.016404 & 0.129966 & 0.249915 & \bfseries 0.293066 \\
(single, cosine) & 0.004966 & \bfseries 0.009726 & 0.003661 & 0.008308 & 0.008689 & \bfseries 0.009726 \\
(complete, cosine) & \bfseries 0.695384 & 0.309572 & 0.420896 & 0.281730 & 0.356601 & 0.309572 \\
(average, cosine) & \bfseries 0.612819 & 0.292957 & 0.273538 & 0.319111 & 0.448297 & 0.292957 \\
(single, manhattan) & 0.006766 & 0.007777 & \bfseries 0.007989 & 0.005402 & 0.003136 & 0.007777 \\
(complete, manhattan) & \bfseries 0.550522 & 0.301899 & 0.030703 & 0.273196 & 0.309888 & 0.301899 \\
(average, manhattan) & 0.021224 & \bfseries 0.291166 & 0.016537 & 0.132257 & 0.269386 & \bfseries 0.291166 \\
(single, l1) & 0.006766 & 0.007777 & \bfseries 0.007989 & 0.005402 & 0.003136 & 0.007777 \\
(complete, l1) & \bfseries 0.550522 & 0.301899 & 0.030703 & 0.273196 & 0.309888 & 0.301899 \\
(average, l1) & 0.021224 & \bfseries 0.291166 & 0.016537 & 0.132257 & 0.269386 & \bfseries 0.291166 \\
(single, l2) & 0.006886 & \bfseries 0.007961 & 0.006723 & 0.004333 & 0.006045 & \bfseries 0.007961 \\
(complete, l2) & \bfseries 0.502632 & 0.296154 & 0.081222 & 0.167146 & 0.299644 & 0.296154 \\
(average, l2) & 0.022881 & \bfseries 0.293066 & 0.016404 & 0.129966 & 0.249915 & \bfseries 0.293066 \\
\bottomrule
\end{tabular}
 
\end{center}
\end{table}

\subsection{DBSCAN}
DBSCAN is the ubiquitous density-based clustering. The mechanism of DBSCAN 
can be summarized as follows~\cite{Schubert2017}.

\begin{enumerate}
 \item Find points within $\epsilon$ distance of every point and identify the
    core points which are points with more than a minimum number of 
    points within distance $\epsilon$.
 \item Determine the connected components of core points on the neighbor graph, 
   excluding all non-core points.
 \item Assign each non-core point to a nearby cluster within an $\epsilon$ 
   distance neighbor, consider it a noisy point.
\end{enumerate}

We experimented with different metric types including euclidean, cosine, and 
manhattan, as well as different values to consider points as neighbors of a 
core point. Table~\ref{tab:dbscan_score} provides the adjusted random score for 
the different experimental configurations for DBSCAN.

\begin{table}[ht]
\begin{center}
  \caption{The adjusted random score of DBSCAN for different pooling strategies.}
  \label{tab:dbscan_score}
  \begin{tabular}{lrrrrrr}
\toprule
 & $X_{subs} $ & $X_{add} $ & $X_{abs} $ & $X_{min} $ & $X_{max} $ & $X_{mean} $ \\
\midrule
(euclidan, 050) & \bfseries 0.028191 & 0.028177 & 0.028177 & 0.028177 & 0.028177 & 0.028177 \\
(cosine, 025) & \bfseries 0.325427 & 0.217478 & 0.014572 & 0.243042 & 0.266950 & 0.217478 \\
(cosine, 030) & \bfseries 0.627861 & 0.043688 & 0.000000 & 0.039113 & 0.261000 & 0.043688 \\
(cosine, 050) & \bfseries 0.512095 & 0.006298 & 0.000000 & 0.003168 & 0.001254 & 0.006298 \\
(manhattan, 050) & \bfseries 0.028191 & 0.028177 & 0.028177 & 0.028177 & 0.028177 & 0.028177 \\
\bottomrule
\end{tabular}
 
\end{center}
\end{table}

\section{Discussion}

\subsection{Results}
Our results suggest that the subtraction pooling strategy might be the best 
operation for word embedding-based word relationship representation, compared
to the five others, experimented with, in the investigation. This finding 
supports the assumption of using the subtraction operator in word vector-based 
analogy tasks~\cite{Mikolov2013b}. Table~\ref{tab:highest_mean} gives the 
model configuration with the highest average score on each pooling strategy
for different clustering models.

\begin{table}[ht]
\begin{center}
  \caption{Configurations with the highest score average from the different 
        clustering models experimented with.}
  \label{tab:highest_mean}
  \begin{tabular}{lrrrrrr}
\toprule
 & $X_{subs} $ & $X_{add} $ & $X_{abs} $ & $X_{min} $ & $X_{max} $ & $X_{mean} $ \\
\midrule
kmean & \bfseries 0.792549 & 0.363478 & 0.296212 & 0.334140 & 0.313051 & 0.363478 \\
gmm & \bfseries 0.853806 & 0.293151 & 0.351499 & 0.329660 & 0.228340 & 0.293151 \\
agglomerative: complete, cosine & \bfseries 0.695384 & 0.309572 & 0.420896 & 0.281730 & 0.356601 & 0.309572 \\
dbscan: cosine, 025 & \bfseries 0.325427 & 0.217478 & 0.014572 & 0.243042 & 0.266950 & 0.217478 \\
\bottomrule
\end{tabular}

\end{center}
\end{table}

In addition, the k-mean centroid-based clustering 
algorithm yields the highest scores for 3 of of 6 pooling strategies. 
The best score comes from clustering relation vectors from 
the subtraction pooling strategy using the Gaussian Mixture clustering model.
Although the Gaussian Mixture Model is a distribution-based clustering mechanism
it is built on top of the centroid-based k-means. This reinforces the superiority
of centroid-based clustering in our experimental setup, followed by the 
agglomerative hierarchical-based clustering configured with complete linkage and 
cosine similarity.


\subsection{Application}
The ability to group representation of similar relationships between pairs of 
words especially named entities. Point toward an unsupervised approach to 
categorize relationships among words including named entities. This can in turn 
be used as link categorization when building knowledge graphs. Thus alleviating 
in some cases the need for manual data labeling for the type of links between 
nodes.
It can go as far as providing data for graph learning models such as 
graph convolutional neural network that aims at including links type in addition
to nodes type in their learning process.

\subsection{Future Work}
This work gives two pointers for further steps. One is extending the 
experimentation, with more word relationship representation strategy and 
clustering models. Additional relationship representation can include 
learned one by using an autoencoder on the pair of word vectors.   
Another direction is to derive a formal explanation of the current findings,
of this exploratory study.

\section{Conclusion}
This study explores possibilities for a word embedding based word to word 
relationship representation and their clustering ability in regards to grouping 
similar relationships. 
Different relationship representations are obtained by applying basic operations
on the coordinate of pairs of vectors. The subtraction pooling strategy and 
centroid-based clustering models tend to give better results in our exploratory
setup. Further work might extend the exploration or provide a formal explanation
of the findings.

\bibliographystyle{splncs04}
\bibliography{arxiv2305}
\end{document}